\documentclass[10pt,twocolumn,letterpaper]{article}

\usepackage{cvpr}
\usepackage{times}
\usepackage{epsfig}
\usepackage{graphicx}
\usepackage{amsmath}
\usepackage{amssymb}

\usepackage[pagebackref=true,breaklinks=true,letterpaper=true,colorlinks,bookmarks=false]{hyperref}

\cvprfinalcopy 

\ifcvprfinal\fi
\begin{document}

\title{Unsupervised Learning of Monocular Depth Estimation with Bundle Adjustment, Super-Resolution and Clip Loss}

\author{Lipu Zhou$^1$, Jiamin Ye$^2$, Montiel Abello$^1$, Shengze Wang$^1$, Michael Kaess$^1$\\
$^1$Carnegie Mellon University, $^2$Chinese Academy of Sciences\\
{\tt\small $^1$\{lipuz,mabello,shengzew,kaess\}@andrew.cmu.edu, $^2$yejiamin@iet.cn}
}

\maketitle

\begin{abstract}
We present a novel unsupervised learning framework for single view depth estimation using monocular videos. It is well known in 3D vision that enlarging the baseline can increase the depth estimation accuracy, and jointly optimizing a set of camera poses and landmarks is essential. In previous monocular unsupervised learning frameworks, only part of the photometric and geometric constraints within a sequence are used as supervisory signals. This may result in a short baseline and overfitting.  Besides, previous works generally estimate a low resolution depth from a low resolution impute image. The low resolution depth is then interpolated to recover the original resolution. This strategy may generate large errors on object boundaries, as the depth of background and foreground are mixed to yield the high resolution depth. In this paper, we introduce a bundle adjustment framework and a super-resolution network to solve the above two problems. In bundle adjustment, depths and poses of an image sequence are jointly optimized, which increases the baseline by establishing the relationship between farther frames. The super resolution network learns to estimate a high resolution depth from a low resolution image. Additionally, we introduce the clip loss to deal with moving objects and occlusion.  Experimental results on the KITTI dataset show that the proposed algorithm outperforms the state-of-the-art unsupervised methods using monocular sequences, and achieves comparable or even better result compared to  unsupervised methods using stereo sequences.
\end{abstract}

\section{Introduction}

Predicting depth from a single image is a challenging task and has many applications in 3D vision and robotics, such as autonomous driving, planning, obstacle avoidance and Simultaneous localization and mapping (SLAM). This task is different from the traditional multiple view reconstruction, which uses a set of images of a scene to recover the 3D information, mainly considering the appearance matching and geometric constraints among these images. Due to its importance, much effort has been made in this task. As in other computer vision tasks, deep learning approaches have achieved great
success. Early works \cite{li2015depth,liu2015deep,wang2015towards,liu2016learning} formulated single image depth prediction as a supervised learning problem. The difficulty of the supervised method lies in the lack of ground truth depth information. Recent works \cite{godard2017unsupervised,zhou2017unsupervised,godard2017unsupervised,zou2018df,zhan2018unsupervised,wang2018learning,yang2018unsupervised,aleotti2018generative} show that view synthesis can be an effective supervisory signal to train the neural network. This removes the requirement of ground truth depth supervisory labels, and makes unsupervised learning for depth estimation possible. 
\begin{figure}
	\centering
	\includegraphics [width=0.5\textwidth ] {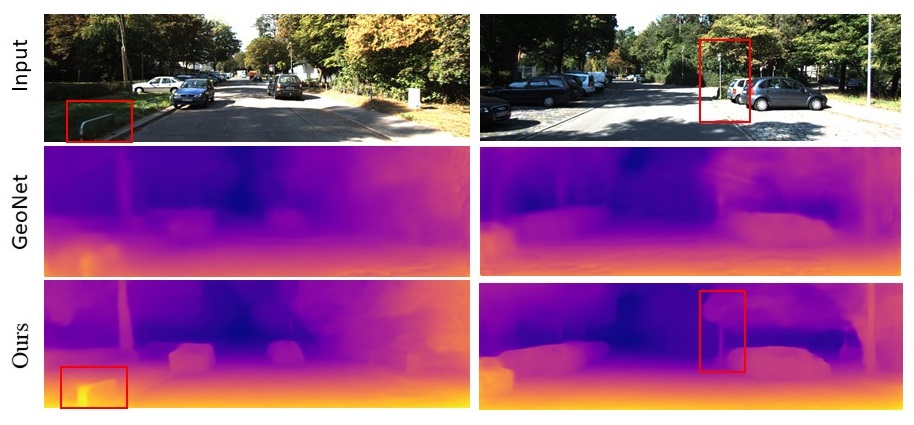}	
	\caption{Experimental results of our algorithm and GeoNet \cite{yin2018geonet} on the KITTI dataset \cite{KITTI}. Our algorithm introduces a super-resolution network to produce a high resolution depth map from a low resolution input. Compared to GeoNet, the depth maps of our algorithm have clear boundaries, and contain more details as shown by the red rectangle. }
	\label{fig:introduction}
\end{figure}

In the literature, stereo and monocular videos are explored to train the network \cite{godard2017unsupervised,zhou2017unsupervised,godard2017unsupervised,zou2018df,zhan2018unsupervised,wang2018learning,yang2018unsupervised,aleotti2018generative}. Compared to stereo, the monocular video is a broader training source and is easier to capture. However, training  on monocular video is more challenging, due to the unknown camera motion, moving objects, and varying lighting conditions. Although current works \cite{zhou2017unsupervised,wang2018learning,zou2018df,yang2018unsupervised,yang2018every}  using monocular videos have showed promising results, there still exists significant gap between the results obtained by stereo and monocular training strategies. This paper focuses on unsupervised learning using monocular videos and seeks to reduce this gap. The contributions of the paper are as follows:

\textbf{Bundle Adjustment Framework} \quad We introduce a bundle adjustment framework to train the network. It is well known that a large baseline is essential for accurate depth estimation. In SLAM or VO systems \cite{mur2017orb,engel2014lsd,engel2018direct}, bundle adjustment is used to jointly optimize a set of camera poses and landmarks, which increases the baseline of a moving camera. Our bundle adjustment framework jointly optimize depths  and camera poses within a sequence. Compared to previous works \cite{zhou2017unsupervised,yin2018geonet,zou2018df,wang2018learning,mahjourian2018unsupervised} that use consecutive frames to generate constraints, our method increases the baseline and introduces more constraints. 

\textbf{Super-resolution Network} \quad Motivated by super-resolution of image \cite{lai2018fast,wang2018recovering}, we introduce a super-resolution network to generate a high resolution depth map from a low resolution input. In previous works, the image is downsampled and fed into the network. The network produces a low resolution depth map which has the same resolution as the input image. This low resolution depth map is then upsampled to recover the depth of the original image. The interpolation combines nearby depths, which may result in large errors, especially at the boundaries of objects. We solve this problem by introducing  a super-resolution network which learns to generate a high resolution depth map. 

\textbf{Clip Loss Function} \quad We introduce a clip loss to deal with the  moving object and occlusion. The supervisory signal of the unsupervised learning framework comes from view synthesis. The resulting loss function is under the assumption of static scene and photometric consistency.  One challenge of learning depth from monocular videos lies in the moving object and occlusion, which will violate the static assumption. The large errors coming from these regions will degrade the performance. This paper introduces a clip loss function to deal with this problem. During training, errors higher than a certain percentile will be capped. They will generate zero gradients and will not impact on training. 

\subsection{Related Work}
Estimating depth from a single image is a challenging task. This differs from the traditional structure-from-motion (SfM) \cite{schonberger2016structure} or multi-view stereo (MVS)  \cite{schonberger2016pixelwise}, where  multiple images are used to recover the depth. A large number of learning based approaches, including both supervised and unsupervised approaches, have been proposed to address the single image depth estimation problem, and great progress has been made in this task.

\noindent
\textbf{Supervised Depth Estimation}   Most supervised approaches formulate the depth estimation problem as a supervised regression problem. In early works \cite{saxena2006learning,saxena2009make3d}, Markov random field(MRF) with hand-crafted features was trained to estimate the depth. Liu \etal \cite{liu2010single} introduced semantic labels into the MRF learning. Ladick\'y \etal \cite{ladicky2014pulling} showed that combining semantic labeling and depth estimation can benefit each other.  Karsch \etal \cite{karsch2014depth} adopted non-parametric sampling for pose estimation. To avoid feature engineering, supervised learning using deep neural networks has been explored.   Eigen \etal \cite{EigenDepth} presented a multi-scale  deep convolutional neural network(CNN) to predict the depth, which is later extended for depth prediction, surface normal estimation, and semantic labeling \cite{eigen2015predicting}. Due to the promising results demonstrated by this approach, various deep network structures have been explored to further improve performance, such as combining CNNs with conditional random field(CRF) \cite{li2015depth,liu2015deep,wang2015towards,liu2016learning}, using fully convolutional residual networks with reverse Huber loss \cite{laina2016deeper}, formulating the depth estimation problem as a pixel-wise classification task \cite{cao2017estimating}, or jointly learning depth and camera motion from two unconstrained images \cite{ummenhofer2017demon}. Recently, Cheng \etal \cite{cheng2018depth} proposed a convolutional spatial propagation network to learn the affinity matrix for depth prediction. Besides CNNs, recurrent neural networks (RNN) is also explored to yield spatio-temporally accurate monocular depth prediction \cite{kumar2018depthnet}.     In these works, depth sensors are used to produce supervisory signs.  The depth information from RGB-D sensors is noisy, and has limited range (generally for indoor scenarios). On the other hand, the depth measurements from LiDARs are sparse and need an accurate GPS/IMU device to register the LiDAR scan \cite{Geiger2012CVPR}. To reduce the requirement of supervisory depth, some works \cite{zoran2015learning,chen2016single} showed that relative depth can be used to  learn the metric depth. Kuznietsov \etal \cite{kuznietsov2017semi} presented a semi-supervised algorithm which combines sparse ground-truth depth and photometric consistency as supervisory signs.      Li \etal \cite{li2018megadepth} adopted structure-from-motion (SfM)  and multi-view stereo (MVS) technology to generate the supervisory 3D information.  This method is not applicable to scenarios where SfM or MVS  fails to work. Ground truth depth  is still required to adapt the pretrained model to a specific application. In recent work \cite{atapour2018real}, synthetic data with perfect depth were used to train a depth estimation network, and an image style transfer network was trained to convert a real image into the synthetic domain, so that depth could be estimated from real images.

\noindent
\textbf{Unsupervised Depth Estimation} The Photometric consistency assumption for nearby frames gives a way to avoid the requirement of ground truth depth at training time. Although various costs are proposed for unsuperivsed learning, view synthesis \cite{flynn2016deepstereo,xie2016deep3d}  is critical in generating self-supervisory signals for unsupervised learning of depth estimation. Specifically, a source and a target image pair are considered at training time. The network yields the depth of the source image, which together with the target image and the pose between the image pair is used to synthesize the source image. The training is conducted by minimizing the  error between the real source image and the synthesized one.  According to the type of training images, unsupervised approaches can be divided into two categories. 

The first category considers learning depth from  \textbf{stereo sequences}.   The left and right images and the known pose of the stereo camera rig form a self-supervisory loop to train the network. Garg \etal \cite{garg2016unsupervised} first applied this self-supervised methodology on stereo image pairs. They used the Taylor expansion to approximate the cost function for gradient computation, which may result in a sub-optimal objective. To solve this problem, Godard \etal \cite{godard2017unsupervised} applied the spatial transformer network \cite{jaderberg2015spatial} to yield a differentiable reconstruction cost function. They also enforced geometric constraints during the training by introducing the left-right disparity  consistency loss. Poggi \etal \cite{poggi2018learning} extended Godard's work \cite{garg2016unsupervised} to trinocular camera system. They introduced an interleaved training procedure to adapt the trinocular network to binocular input.  Except for the above  photometric reconstruction error of the left and right image pair, the temporal  photometric and deep feature reconstruction errors were also exploited to improve the performance in \cite{zhan2018unsupervised}. Recent works \cite{pilzer2018unsupervised,aleotti2018generative} showed that  generative adversarial network (GAN) \cite{radford2015unsupervised} paradigm, and supervision from stereo matching network trained by the
depth from synthetic data \cite{guo2018learning} can benefit unsupervised depth learning.

The second category focuses on learning depth from \textbf{monocular sequences}. Compared to the above case,  this is a more challenging problem, as the camera pose is unknown. Zhou \etal \cite{zhou2017unsupervised} and Vijayanarasimhan \etal \cite{vijayanarasimhan2017sfm} showed that it is capable of  learning depth prediction and pose estimation at the same time using the supervisory signs from view reconstruction error and spatial smoothness cost. The pose estimation network removes the requirement of stereo training samples. Several recent works explored introducing different constraints during training. The consistency between normal and depth was utilized in \cite{yang2018unsupervised}. The 3D point cloud alignment loss was introduced in \cite{mahjourian2018unsupervised}.  Depth and optical flow prediction are related tasks. Recent works \cite{yin2018geonet,zou2018df} show that jointly learning depth and optical flow can benefit each other.  Motivated by current direct visual odometry (DVO) technologies \cite{engel2014lsd,engel2018direct}, Wang \etal \cite{wang2018learning} introduced a differentiable DVO (DDVO) module to replace the previous pose estimation network. They also presented a simple depth normalization strategy to address the scale sensitivity problem cased by the generally used depth regularization term. Godard \etal \cite{godard2018digging} presented several ways to improve the depth estimation, including using a pretrained encoder, sharing lower layers between pose and depth estimator, and training each lower resolution depth map by upsampling them to the input image resolution.

Moving objects are another problem for  training using monocular sequences. The stereo camera pair capture objects at the same, so the scene satisfies the rigid transformation and the moving object is not problematic.    Vijayanarasimhan \etal \cite{vijayanarasimhan2017sfm} dealt with this problem by introducing an object motion model and an object mask. This method requires knowledge of the number of moving objects in the scene, which is difficult to be estimated.  Zhou \etal \cite{zhou2017unsupervised} proposed to an explanation mask to get rid of regions undergoing motion and occlusion. However, they later found that this reduced performance. Yin \etal \cite{yin2018geonet} introduced a ResFlowNet to learn the residual non-rigid flow which is caused by the moving object. The forward-backward consistency is used in \cite{zou2018df,yin2018geonet} to deal with moving objects and occlusions.

\section{Our Approach}
Our framework includes two networks for depth and pose estimation, respectively, as demonstrated in Fig.~\ref{fig:ba} . The depth network  produces a depth map for each image within a sequence. The pose network takes two consecutive images as input, and predicts the relative pose between them. As it sequentially slides through the whole sequence, it estimates the relative pose of each image pairs. This differs from the previous works \cite{zhou2017unsupervised,wang2018learning} which stack the whole sequence as  the input and estimates the poses relative to the center image. This section details our unsupervised framework.

\subsection{Bundle Adjustment Process}

\begin{figure*}
	\centering
	\includegraphics [width=16.9cm ] {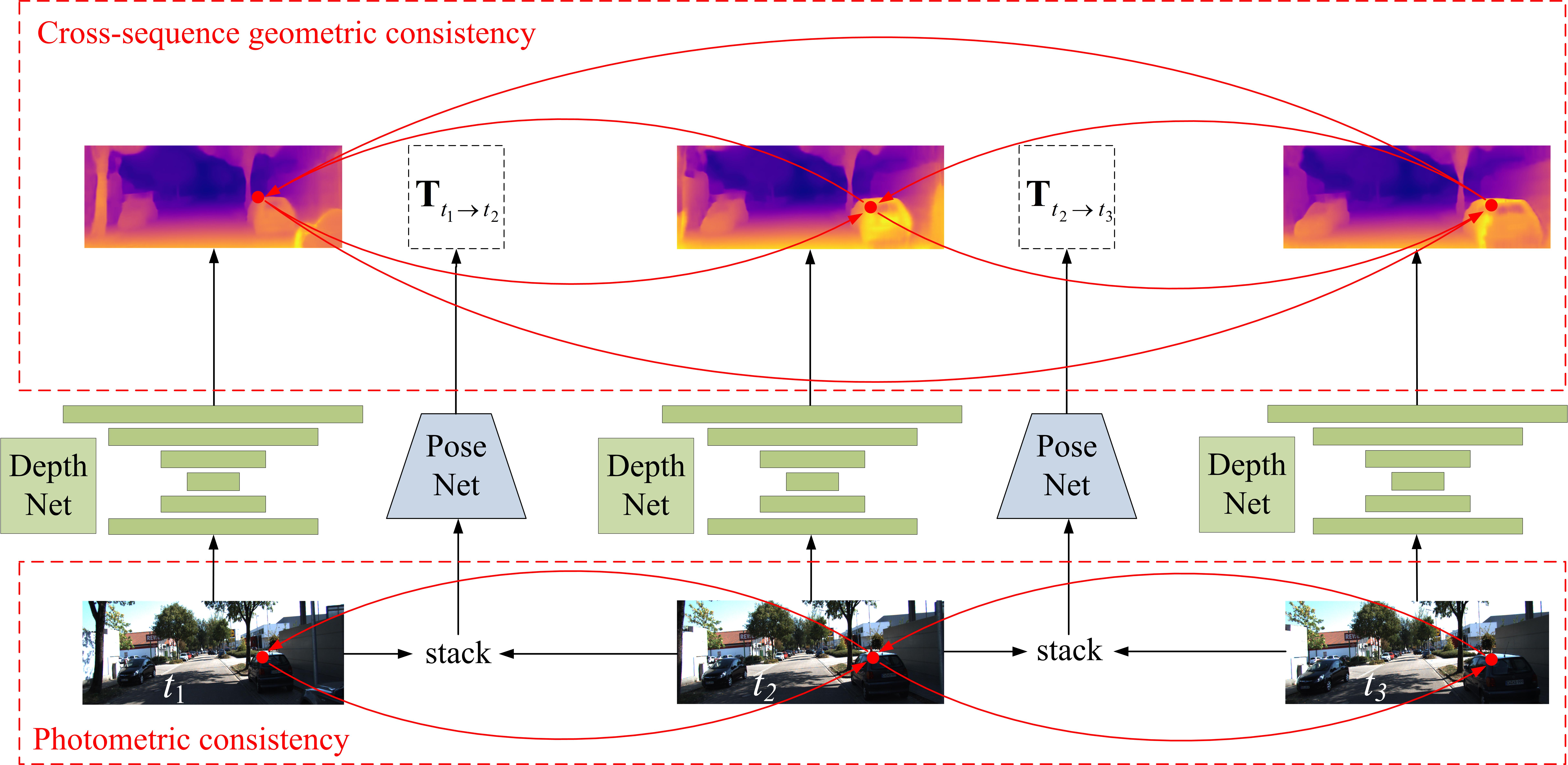}	
	\caption{Our bundle adjustment unsupervised learning framework. In the traditional SLAM or VO system, camera poses and landmarks are jointly optimized in bundle adjustment. Similarly,  our algorithm jointly optimizes depths and camera poses by using  photometric consistency loss and cross-sequence geometric consistency loss.   Arcs represent view synthesis. The cross-sequence connections increase the baseline and  generate more constraints than previous works \cite{zhou2017unsupervised,vijayanarasimhan2017sfm,yang2018unsupervised,yin2018geonet,zou2018df,wang2018learning,mahjourian2018unsupervised}. We do not introduce corss-sequence connections to images, as the lighting condition may vary for distant frames, but the geometric constraints should always maintain.}
	\label{fig:ba}
\end{figure*}

Our network is trained by the supervisory signs from a bundle adjustment process, as demonstrated in Fig. \ref{fig:ba}. Bundle adjustment is an essential component to yield accurate results in traditional SLAM or VO systems \cite{mur2017orb,engel2014lsd,engel2018direct}. A large baseline is important in achieving accurate depth estimation. Bundle adjustment gives an effective way to increase the baseline of a moving camera. Motivated by this, we formulated our unsupervised training in the bundle adjustment manner. In traditional monocular SLAM or VO systems, landmarks are tracked by a certain descriptor \cite{mur2017orb} or photometric consistency \cite{engel2014lsd,engel2018direct} in nearby frames.   Bundle adjustment is then performed to jointly optimize the camera poses and the landmarks. Similarly, in our framework, photometric consistency is used to track each pixel frame by frame, depth consistency is exploited to establish the cross-sequence constraints on  camera poses and depths. Besides the  forward motion, we also consider a backward motion which reverses the sequence during training.   Compared to  previous works \cite{zhou2017unsupervised,vijayanarasimhan2017sfm,yang2018unsupervised,yin2018geonet,zou2018df,wang2018learning,mahjourian2018unsupervised}, our bundle adjustment process yields a larger baseline and more constraints to avoid overfitting.

Our bundle adjustment process is differentiable which leads to training the network end-to-end. It uses N-frame training snippets  $S = \{ {I_1},{I_2}, \cdots ,{I_N}\} $ from video sequences as inputs. The bundle adjustment process over the training snippet $S$   drives the depth network and the pose network to minimize photometric cost, geometric cost and local smoothness cost detailed below.

\subsubsection{Image Reconstruction Loss}

We first consider the image reconstruction error from 2 consecutive images ${{I}_{t}}$ and ${{I}_{t+1}}$. Given the estimated depth map ${{\hat{D}}_{t}}$ from ${{I}_{t}}$ and the estimated pose ${{\hat{T}}_{t\to t+1}}$ between ${{I}_{t}}$ and ${{I}_{t+1}}$, we can map a homogeneous pixel ${{p}_{t}}\in {{I}_{t}}$ onto a homogeneous pixel ${{\hat{p}}_{t+1}}\in {{I}_{t+1}}$ as
\begin{equation} \label{equ:prejection}
{\hat p_{t + 1}} \sim K{\hat T_{t \to t + 1}}{\hat D_t}\left( {{p_t}} \right){K^{ - 1}}{p_t}
\end{equation}
where $K$ is the camera intrinsic parameter. Since ${{\hat{p}}_{t+1}}$ is with continuous coordinates, we apply the differentiable spatial transformer network introduced by \cite{jaderberg2015spatial} to calculate the value of ${{I}_{t+1}}\left( {{{\hat{p}}}_{t+1}} \right)$. Specifically, ${{I}_{t+1}}\left( {{{\hat{p}}}_{t+1}} \right)$ is calculated by the bilinear interpolation using the depths of the 4 neighbors (top-left $\hat{p}_{t+n}^{tl}$, top-right $\hat{p}_{t+n}^{tr}$, bottom-left $\hat{p}_{t+n}^{bl}$, and bottom-right $\hat{p}_{t+n}^{br}$) around ${{\hat{p}}_{t+n}}$. Using this method, we can reconstruct ${{I}_{t}}$ by ${{I}_{t+1}}$ and ${{\hat{T}}_{t\to t+1}}$ as
\begin{equation} \label{equ:view_synthesis}
\hat I\left( {{p_t}} \right) = {I_{t + 1}}\left( {{{\hat p}_{t + 1}}} \right) = \sum\limits_{i \in \{ t,b\} ,j \in \{ l,r\} } {{w^{ij}}I\left( {\hat p_{t + 1}^{ij}} \right)}  
\end{equation}
where ${{w}^{ij}}$ is the coefficient of the bilinear interpolation.   Assuming a static scene, no occlusion, and constant lighting conditions, ${{\hat{I}}_{t}}$ is expected to be the same as ${{I}_{t}}$. As some pixels of ${{I}_{t}}$ may not be visible in ${{I}_{t+1}}$, we use the mask ${{M}_{t}}\left( {{p}_{t}} \right)$ proposed in \cite{mahjourian2018unsupervised} to get rid of these invisible pixels. We formulate the image reconstruction error between ${{\hat{I}}_{t}}$ and ${{I}_{t}}$ as
\begin{equation} \label{equ:lre}
	\begin{array}{l}
	{L_{re}} = \sum\limits_{t = 1}^{N - 1} {\sum\limits_{{p_t}} {{M_t}({p_t})L_{re}^t\left( {{p_t}} \right)} } ,\\
	L_{re}^t\left( {{p_t}} \right) = \alpha \frac{{1 - {\rm{SSIM(}}{{\hat I}_t}({p_t}),{I_t}({p_t}){\rm{)}}}}{2} + (1 - \alpha )\left| {{{\hat I}_t}({p_t}) - {I_t}({p_t})} \right|
	\end{array}
\end{equation}
where $|\cdot |$ represents the absolute value, and SSIM represents the structural similarity index \cite{wang2004image} and $\alpha $ is set to $0.85$ as in the previous works \cite{godard2017unsupervised,wang2018learning}.

\subsubsection{Cross-sequence Geometric Consistency Loss}
The depth network can predict the depth of each image in $S$. The depth of a 3D point estimated from different images should be consistent. This can be used to establish the connections among images in $S$.

Assume ${{D}_{t}}$ and ${{D}_{t+n}}$, $n\ge 1$ are the depth maps from image ${{I}_{t}}$ and ${{I}_{t+n}}$, respectively. For each ${{p}_{t}}\in {{I}_{t}}$, we can use (\ref{equ:prejection}) to estimate the corresponding ${{\hat{p}}_{t+n}}\in {{I}_{t+n}}$. As ${{\hat{p}}_{t+n}}$ is of continuous coordinates, we estimate the depth of ${{\hat{p}}_{t+n}}$ by the bilinear interpolation as the image reconstruction (\ref{equ:view_synthesis}). That is to say we can estimate the  depth of ${{p}_{t}}$ in frame $t+n$ by ${{D}_{t+n}}$. We denote the depth map generated in this way as ${{\hat{D}}_{t\to t+n}}\left( p_t \right)$. On the other hand, we can transform the point cloud in frame $t$ to frame $t+n$ using 
\begin{equation} \label{equ:transfrom}
	{P_{t \to t + n}} = {\hat T_{t \to t + n}}{\hat D_t}\left( {{p_t}} \right){K^{ - 1}}{p_t}
\end{equation} 
where ${{\hat{T}}_{t\to t+n}}$ is the estimated pose from ${{I}_{t}}$ to ${{I}_{t+n}}$. Then the depth of ${p_t}$ in frame $t+n$ is the z-coordinate of ${{P}_{t\to t+n}}$. Denote the depth generated from (\ref{equ:transfrom}) as ${{D}_{t\to t+n}}\left( {{p}_{t}} \right)$.  Ideally, ${{D}_{t\to t+n}}\left( {{p}_{t}} \right)$ and ${{\hat{D}}_{t\to t+n}}\left( p \right)$ should be equal. Therefore, we define the following depth consistency loss
\begin{equation} \label{equ:gcl}
	{L_{dc}} = \sum\limits_{t = 1}^{N - 1} {\sum\limits_{n = 1}^{N - t} {\sum\limits_{{p_t}} {{M_{t \to t + n}}({p_t})} \left| {{D_{t \to t + n}}({p_t}) - {{\hat D}_{t \to t + n}}({p_t})} \right|} }
\end{equation}
${{L}_{dc}}$ establishes cross-sequence constraints, as illustrated in Fig. \ref{fig:ba}. Our formulation is also different from the point cloud alignment loss in \cite{mahjourian2018unsupervised}. The training in \cite{mahjourian2018unsupervised} is to minimize the residual of two point cloud alignment using an ICP procedure. The ICP procedure may converge to a local minimal solution, which results in suboptimal objective. Our algorithm is also different from \cite{godard2017unsupervised}. In \cite{godard2017unsupervised}, the left and right disparities are all from the left image, and the left-right consistency loss is limited to the stereo image pair. Our depth consistency is across the whole sequence. This differs from the works \cite{vijayanarasimhan2017sfm,zou2018df} that limit the depth consistency error to two frames. Our formulation increases the baseline and introduce more constraints, which improves depth estimation.

\subsubsection{Spatial Smoothness}
The above cost function is not sufficient to estimate the depth in textureless regions. To handle this problem, we adopt the edge-aware smoothness regularization term employed in previous works \cite{godard2017unsupervised,mahjourian2018unsupervised,zou2018df} to encourage local smoothness while allowing sharpness at the edge. As the range of depth is unbounded, we impose the following regularization term on the disparity (inverse depth) map
\begin{equation}{\label{equ:lds}}
{L_{ds}} = \sum\limits_{t = 1}^N {\sum\limits_{i,j} {\left| {{\partial _x}d_{ij}^t} \right|{e^{ - |{\partial _x}I_{ij}^t|}} + \left| {{\partial _y}d_{ij}^t} \right|{e^{ - |{\partial _y}I_{ij}^t|}}} } 
\end{equation}
where ${{\partial }_{x}}$ and ${{\partial }_{x}}$ represent the gradient in x and y direction, respectively.

\subsubsection{Backward Sequence}
In the forward motion sequence, only part of the pixels in current frame can be observed in the next frame, but most pixels in the next frame are generally visible in the current frame. Besides the aforementioned forward motion process, we also reverse the order of the sequence to generate a backward sequence. We construct the cost for the backward sequence in the same manner as the forward sequence. This leads to more constraints to avoid overfitting. During training, we jointly optimize the forward and backward losses. 

\subsection{Clip Loss Function}

The above model assumes a static scene, no occlusion and brightness constancy. The parts of the image that violate the above assumption will generate a large cost, and in turn yield a large gradient, which potentially worsens the performance. We treat these violations as outliers, and present a clipping function to handle them. Assume ${{s}_{i}}$ is the $i$th cost of the entire cost set $\mathbf{S}$, \ie, ${{s}_{i}}\in \mathbf{S}$.  To handle the above problem, we introduce the following robust loss function
\begin{equation} \label{equ:clf}
	\rho \left( {{s_i}} \right) = \min \left( {{s_i},\alpha } \right),\alpha  = p\left( {{\bf{S}},q} \right)
\end{equation}
where  $p\left( {{\bf{S}},q} \right)$ represents the  $qth$ percentile of  $S$. That is to say the cost in  $S$ is clipped at the $qth$ percentile. Costs above the $qth$ percentile will yield zero gradient, and do not affect the training. We apply (\ref{equ:clf}) to the cost functions (\ref{equ:lre}) and (\ref{equ:gcl})  introduced above.

\subsection{Super-resolution Network}

In the previous works, the network takes a down-sampled image as input and provides the same resolution depth map. This depth map is then scaled to the original resolution. The drawback of this process is that it may generate large errors at object boundaries in the high-resolution depth map, as the interpolation simply combines depth values of background and foreground. To solve this problem, we introduce a super-resolution network. The super-resolution network takes the low-resolution image and the output of the original decoder as input, and yields a 2 times higher resolution depth map. From Fig.~\ref{fig:introduction} and \ref{fig:qualitative_rlt}, we can find that our network can generate clear boundaries and more details of the scene.

\subsection{Total Objective Function }
Our learning objective combines the abovementioned loss functions including forward and backward motion. For the cost function (\ref{equ:lre}) and (\ref{equ:gcl}), we apply (\ref{equ:clf}) to deal with moving objects and occlusion. We adopt multiple-scale loss from coarse to fine to train the network. 4 scales are used in the experiments as previous works \cite{zhou2017unsupervised,godard2017unsupervised}. The final objective function is
\begin{align} \label{equ:objective}
&L = \sum\limits_{s = 1}^4 {\frac{{{L_s}}}{{{2^{s - 1}}}}} ,\\
&{L_s} = \rho ({}^sL_{re}^f) + \rho ({}^sL_{re}^b) + \alpha (\rho ({}^sL_{dc}^f) + \rho ({}^sL_{dc}^b)) + \beta {}^s{L_{ds}} \nonumber
\end{align}

\subsection{Network Architecture}
Here we introduce our network architecture. Our network includes two components, \ie, a single-view depth network and a camera pose network. We adopt the network architecture of \cite{zhou2017unsupervised} as the backbone. 

\textbf{Depth Network} \quad Our depth network takes each image in   as input, and provides a two times larger dense depth map. Our super-resolution layer is mounted on the last convolutional layer of the depth network in \cite{zhou2017unsupervised}. The depth network in \cite{zhou2017unsupervised} is of an encoder-decoder structure with skip connection between encoder and decoder. Th VGG network \cite{simonyan2014very} is used as the encoder. ReLU is adopted as the activation function, except for the inverse depth prediction layer where the sigmoid activation function is used. We also adopt the depth normalization proposed in \cite{wang2018learning} to deal with the scale sensitivity of the depth regularization term. The raw image together with the output of the decoder network are stacked and fed into the super-resolution network. This uses transpose convolution to double the resolution of the input image. Our network uses 4 multi-scale training objectives downsampling from the super-resolution layer. 

\textbf{Pose Network} \quad The pose network receives two consecutive images  as input, and predicts the relative pose between them. The network goes through the sequence. Then the pose between arbitrary images in the sequence   can be easily derived from the outputs of the pose network. This differs from \cite{zhou2017unsupervised,wang2018learning}, where the whole image sequence is fed into the pose network, and the poses between the middle image and others are predicted.

\section{Experiments}
In this section, we extensively evaluate our algorithm through experiments. We compare our algorithm with the state-of-the-art methods in terms of accuracy and cross-dataset generalization ability. In addition, we conduct ablation experiments to show that the proposed bundle adjustment framework, super-resolution network and clip loss function all benefit single view depth prediction. 

\subsection{Training Details}
Our network is implemented in the TensorFlow framework. We employed the Adam \cite{kingma2014adam}  optimizer to minimize the objective function (\ref{equ:objective}) with ${{\beta }_{1}}=0.9$, ${{\beta }_{2}}=0.999$. The model is trained for 20 epochs  with initial learning rate of ${{10}^{-4}}$ for the first 15 epochs, then dropped to ${{10}^{-5}}$ for the last 5 epochs. We tested two input resolutions including $128\times 416$ and  $192\times 640$ with batch size 4 and 3, respectively. The weights in (\ref{equ:objective}) are set as $\alpha =1,\beta =0.01$ throughout all the experiments bellow. For the clip loss function (\ref{equ:clf}), we set the percentile $q=95$ in our experiments. That is to say errors larger than $95\%$ errors 
will be clipped and generate zero gradient. During training, we randomly scale the  image contrast with $[0.8, 1.2]$ and jet the brightness with $ \pm 10$.

\subsection{Monocular Depth Estimation}
We mainly use the KITTI dataset \cite{KITTI} to evaluate our monocular depth estimation algorithm. As in previous works, we set the length of training sequences to 3.  The dataset is split as \cite{EigenDepth}, which generates 40K training samples, 4k evaluation samples, and 697 test samples. We follow the general metrics of the depth estimation in related works \cite{zhou2017unsupervised,wang2018learning}. The performance is assessed by absolute relative difference, square related difference, RMSE and log RMSE. We consider two input resolutions, \ie, $128\times 416$ and $192\times 640$, which result in $256\times 832$ and $384\times 1280$ depth map, respectively. We use the postfix LR to refer to the lower resolution input. 

Table \ref{table:kitti} lists the results of our algorithm and previous works. Our algorithm with  $192\times 640$  input generates better results than $128\times 416$   input.  The depth map resolution  $384\times 1280$  from  $192\times 640$  input better approximates the original image resolution in KITTI, so it suffers less from  the  interpolation. For training using the KITTI dataset, our algorithm significantly outperforms previous unsupervised algorithms using monocular videos. Our algorithm yields comparable or even better results than the state-of-the-art algorithms using stereo sequences to train the network.  For the stereo training strategy, only the work \cite{poggi2018learning,mehta2018structured}  using ResNet \cite{he2016deep} as encoder generates better metrics on Abs Rel and Sq Rel than our algorithm. Our algorithm uses VGG as encoder, and outperforms the VGG version of \cite{poggi2018learning}. Our algorithm fills the gap between  monocular and stereo training strategies.

\begin{figure*}
	\centering
	\includegraphics [width=17cm ] {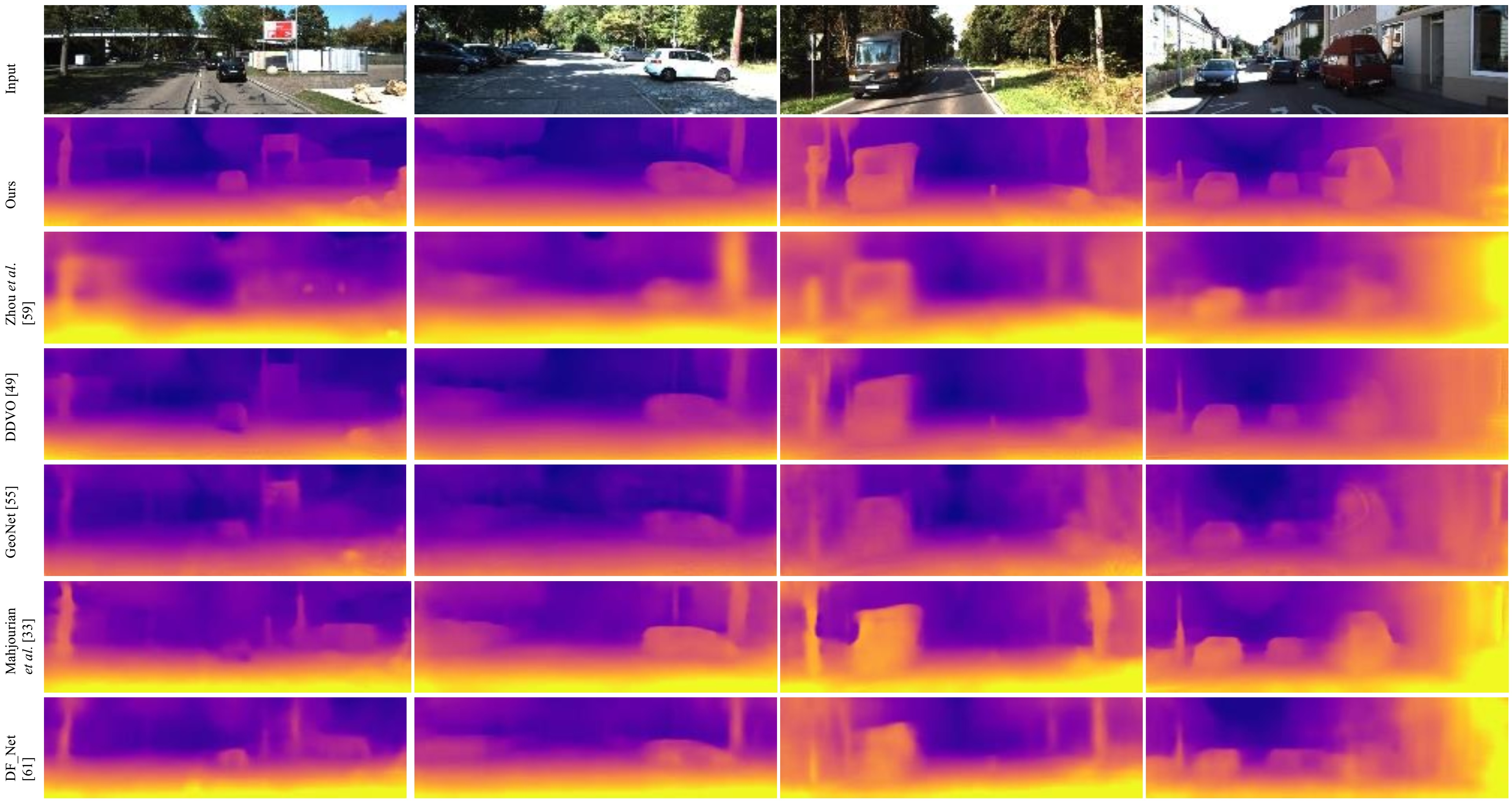}	
	\caption{Qualitative results on test images from the KITTI Eigen split.} 
	\label{fig:qualitative_rlt}
\end{figure*}

Fig.~\ref{fig:qualitative_rlt} provides some qualitative results on the KITTI dataset. The previous works directly upsample the depth image, so their results are generally blurry, and their depth maps lose some details of the image.  

We also consider using CityScape dataset \cite{cityscapes} to pretrain the model, then tuning on the KITTI dataset. We find that pretraining on the CityScape dataset significantly benefits the stereo training  strategy, but only slightly improve the result of the monocular training strategy. The work \cite{yin2018geonet} even reported worse results. One possible reason is that frame rates of the video and the motion patterns of the vehicle, such as speed, differ during recording the two datasets. Therefore, the pose network trained on the CityScape dataset will not work well on the KITTI dataset. This will in turn impact on the training of the depth network. Methods based on stereo do not require the pose network, so the extra data can improve the performance.

\begin{table*}
	\small
	\begin{center}
		\begin{tabular}{ c|c|c|c|c|c|c|c}
			\hline
			Method    & Abs Rel & Sq Rel & RMSE & RMSE log & $\delta  < 1.25$ & $\delta  < 1.25^2$ &  $\delta  < 1.25^3$\\
			\hline\hline
			Ours                            & 0.143 & 1.104 & 5.370 & 0.219 & 0.824 & 0.937& 0.974  \\
			Ours w/o CLF                    	& 0.146 & 1.172 & 5.435 & 0.222 & 0.820 & 0.934 & 0.972 \\
			Ours w/o CLF and CSGCL            & 0.150 & 1.272 & 5.594 & 0.228 & 0.817 & 0.934 & 0.970 \\ 
			Ours w/o CLF, SR and CSGCL        & 0.154 & 1.341 & 5.824 & 0.234 & 0.802 & 0.932 & 0.971 \\
			\hline
		\end{tabular}
	\end{center}
	\caption{ Ablation study of our algorithm on the KITTI dataset \cite{KITTI}  using the split of Eigen \etal \cite{EigenDepth}. Depths are  capped at 80m. CLF, SR and CSGCL represent clip loss function, super-resolution and cross-sequence geometric consistency loss, respectively. The input resolution is $128 \times 416$ pixels.} \label{table:ablation} 
\end{table*}

\begin{table}
	\small
	\begin{center}
		\begin{tabular}{@{} l @{}|@{} c @{}|@{}c|@{}c|@{}c|@{} c @{}}
			\hline
			Method    & Supervision & Abs Rel & Sq Rel & RMSE & RMSE log \\
			\hline\hline
			Karsch \etal \cite{karsch2014depth} & Depth & 0.428 & 5.079 & 8.389 & 0.149 \\
			Liu \etal \cite{liu2014discrete} & Depth & 0.475 & 6.562 & 10.05 & 0.165 \\
			Laina \etal \cite{laina2016deeper} & Depth & \textbf{0.204} & \textbf{1.840} &\textbf{5.683} & \textbf{0.084} \\
			\hline
			MegaDepth \cite{li2018megadepth} & SfM+MVS & \textbf{0.298} & - & \textbf{5.49} & \textbf{0.115} \\
			Atapour \etal \cite{atapour2018real} &SS& 0.423 & 9.343 & 9.002 & 0.122 \\
			Godard \etal \cite{godard2017unsupervised} & Pose & 0.544 & 10.94 & 11.76 & 0.193 \\			
			Zhou \etal \cite{zhou2017unsupervised} & No & 0.383 & 5.321 & 10.47 & 0.478 \\
			DDVO \cite{wang2018learning} & No & 0.387 & 4.720 & 8.09 & 0.204 \\
			DF-Net \cite{zou2018df} & No & 0.331 &\textbf{ 2.698} & 6.89 & 0.416 \\
			Ours\_LR     & No & 0.360 & 3.586 & 7.458 & 0.205 \\
			Ours & No & 0.350 & 3.853 & 7.387 & 0.185 \\ 
			\hline
		\end{tabular}
	\end{center}
	\caption{Results on the Make3D dataset \cite{saxena2009make3d}. Our algorithm and \cite{li2018megadepth,atapour2018real,godard2017unsupervised,zhou2017unsupervised,engel2018direct,zou2018df} do not exploit the Make3D data to train the model. We directly apply the model trained on KITTI+Cityscapes to the test dataset of Make3D. Errors are calculated only for pixels in the central image crop whose depths are less than 70 meters. SD represents supervision using synthetic data.} \label{table:make3d}
\end{table}

\begin{table*} 
	\small
	\begin{center}
		\begin{tabular}{l|c|c|c|c|c|c|c|c|c}
			\hline
			Method & Supervision & Dataset  & Abs Rel & Sq Rel & RMSE & RMSE log & $\delta  < 1.25$ & $\delta  < 1.25^2$ &  $\delta  < 1.25^3$\\			
			\hline\hline
			Eigen \cite{EigenDepth} & Depth & K & 0.203 & 1.548 & 6.307 & 0.282 & 0.702 & 0.890 & 0.958 \\
			MegaDepth \cite{li2018megadepth} & SfM+MVS & MD+K & 0.141 & 1.328 & 5.90 & 0.241 & - & - & -  \\
			Kuznietsov \etal \cite{kuznietsov2017semi}  & Depth+Pose & K & 0.113 & \textbf{0.741} & \textbf{4.621} & \textbf{0.189} &  0.862 & 0.960 & \textbf{0.986} \\			
			Atapour \etal \cite{atapour2018real} & SS & S+K & 0.110 & 0.929 & 4.726 & 0.194 & \textbf{0.923} & \textbf{0.967} & 0.984 \\
			Guo \etal \cite{guo2018learning} & SS + Pose & S+K & \textbf{0.105 }& 0.811 & 4.634 & \textbf{0.189} & 0.874 & 0.959 & 0.982	 \\
			\hline				
			Garg \etal \cite{garg2016unsupervised} cap 50m & Pose & K & 0.169 & 1.080 & 5.104 & 0.273 & 0.740 & 0.904 & 0.962 \\
			PyD-Net(200) \cite{poggi2018towards} & Pose & K & 0.153 & 1.363 & 6.030 & 0.252 & 0.789 & 0.918 & 0.963 \\
			Godard \etal \cite{godard2017unsupervised} & Pose & K & 0.148 & 1.344 & 5.927 & 0.247 & 0.803 & 0.922 & 0.964 \\			 
			Pilzer \etal \cite{pilzer2018unsupervised} & Pose & K & 0.152 & 1.388 & 6.016 & 0.247 & 0.789 & 0.918 & 0.965 \\
			Zhan \etal \cite{zhan2018unsupervised} & Pose & K & 0.135 & 1.132 & 5.585 & 0.229 & 0.820 & 0.933 & 0.971 \\
			StrAT ResNet50 \cite{mehta2018structured} & Pose & K &  \textbf{0.128} & 1.019 & 5.403 & 0.227 & 0.827 & 0.935 & 0.971\\ 
			3Net \cite{poggi2018learning} & Pose & K & 0.142 & 1.207 & 5.702 & 0.240 & 0.809 & 0.928 & 0.967 \\
			3Net ResNet50 \cite{poggi2018learning} & Pose & K & 0.129 & \textbf{0.996} & \textbf{5.281} & \textbf{0.223} & \textbf{0.831} & \textbf{ 0.939} & \textbf{0.974} \\
				
			\hline
			Kumar \etal \cite{kumar2018monocular}  & No & K & 0.211 & 1.980 & 6.154 & 0.264 & 0.732 & 0.898 & 0.959 \\
			Zhou \etal \cite{zhou2017unsupervised} & No & K & 0.208 & 1.768 & 6.856 & 0.283 & 0.678 & 0.885 & 0.957 \\ 
			Yang \etal \cite{yang2018unsupervised} & No & K & 0.182 & 1.481 & 6.501 & 0.267 & 0.725 & 0.906 & 0.963\\
			Mahjourian \etal \cite{mahjourian2018unsupervised} & No & K & 0.163 & 1.240 & 6.220 & 0.250 & 0.762 & 0.916 & 0.968\\
			LEGO \cite{LEGO} & No & K & 0.162 & 1.352 & 6.276 & 0.252 & - & - & -	\\	
			GeoNet \cite{yin2018geonet} & No & K & 0.155 & 1.296 & 5.857 & 0.233 & 0.793 & 0.931 & 0.973 \\	
			DDVO \cite{wang2018learning} & No & K & 0.151 & 1.257 & 5.583 & 0.228 & 0.810 & 0.936 & 0.974\\
			DF-Net \cite{zou2018df} & No & K & 0.150 & 1.124 & 5.507 & 0.223 & 0.806 & 0.933 & 0.973\\
			\textbf{Ours\_LR}     & No & K & 0.143 & 1.104 & 5.370 &  0.219 & 0.824 & 0.937 & 0.974 \\
			\textbf{Ours} & No & K & \textbf{0.139} & \textbf{1.057} & \textbf{5.213} &  \textbf{0.214} & \textbf{0.831} & \textbf{0.940} & \textbf{0.975} \\
			\hline \hline
			
			PyD-Net(200) \cite{poggi2018towards} & Pose & CS+K & 0.146 & 1.291 & 5.907 & 0.245 & 0.801 & 0.926 & 0.967 \\
			Godard \etal \cite{godard2017unsupervised} & Pose & CS+K & 0.124 & 1.076 & 5.311 &  0.219 & 0.847 & 0.942 & 0.973 \\
			MonoGAN \cite{aleotti2018generative} & Pose & CS+K & 0.124 & 1.055 & 5.289 & 0.220 & 0.847 & 0.942 & 0.973 \\
			3Net \cite{poggi2018learning} & Pose & CS+K & 0.117 & 0.905 & 4.982 & 0.210 & 0.856 & 0.948 & 0.976 \\
			3Net ResNet50 \cite{poggi2018learning} & Pose & CS+K & \textbf{0.113} & \textbf{0.885} & \textbf{4.898} & \textbf{0.204} & \textbf{0.862} & \textbf{0.950} & \textbf{0.977} \\
			\hline
			Zhou \etal \cite{zhou2017unsupervised} & No & CS+K  & 0.198 & 1.836 & 6.565 & 0.275 & 0.718 & 0.901 & 0.960 \\
			Mahjourian \etal \cite{mahjourian2018unsupervised} & No & CS+K & 0.159 & 1.231 & 5.912 & 0.243 & 0.784 & 0.923 & 0.970 \\
			LEGO \cite{LEGO} & No & CS+K & 0.159 & 1.345 & 6.254 & 0.247 & - & - & - \\
			GeoNet \cite{yin2018geonet} & No & CS+K & 0.153 & 1.328 & 5.737 & 0.232 & 0.802 & 0.934 & 0.972 \\			
			DDVO \cite{wang2018learning} & No & CS+K & 0.148 & 1.187 & 5.496 & 0.226 & 0.812 & 0.938 & 0.975 \\	
			DF-Net \cite{zou2018df} & No & CS+K & 0.146 & 1.182 & 5.215 & \textbf{0.213} & 0.818 & \textbf{0.943} & \textbf{0.978} \\	
			\textbf{Ours\_LR}           & No & CS+K & 0.142 & 1.090 & 5.380 & 0.219 & 0.822 & 0.938 & 0.974 \\
			\textbf{Ours}       & No & CS+K & \textbf{0.139} & \textbf{1.043} & \textbf{5.160} & 0.215 & \textbf{0.833} & 0.939 & 0.975 \\
			\hline
		\end{tabular}
	\end{center}
	\caption{Single-view depth prediction results on the KITTI dataset using the split of Eigen \etal \cite{EigenDepth}. The dataset column lists the training dataset. K, CS and S denotes KITTI dataset \cite{KITTI}, Cityscapes dataset \cite{cityscapes} and synthetic data, respectively. For supervision, MegaDepth \cite{li2018megadepth} use depth from SfM and MVS. The works  \cite{atapour2018real,guo2018learning} use synthetic data with perfect ground truth to pretrain the model. We refer this supervision as synthetic supervision (SS).   The table is divided into several sections according to the type of supervision, dataset used to training and capped distance. The best result of each section is in bold. All the results are from the depth capped at 80m, except for the result of Garg \etal \cite{garg2016unsupervised} which is capped at 50m.}\label{table:kitti}
\end{table*}

\subsection{Ablation Study}
We conduct an ablation study to evaluate the importance of different components. We trained a series of models. Each one drops one proposed component.  Table \ref{table:ablation} lists the results. It is obvious that all the proposed components improve the performance. We find that our algorithm without super-resolution, cross-sequence geometric consistency loss and clip loss function gives similar results as \cite{yin2018geonet}, where constraints between consecutive frames are used to construct the loss. Removing clip loss, dynamic objects and occlusions  impacts on the training. Dropping the cross-sequence geometric consistency loss reduces the baseline. Without super-resolution, the bilinear interpolation  introduces large errors at object boundaries.

\section{Generalization Ability}
We use the Make3D dataset \cite{saxena2009make3d} to test the cross dataset generalization ability of the algorithm. We use the model trained on the CityScape and KITTI to the test dataset of Make3D. Table \ref{table:make3d} lists the results. Our algorithm gives competitive results. The depth map resolution in the  Make3D dataset is low ($ 21 \times 305 $).  We have to downsample our high resolution depth map to a lower one. As upsampling, downsampling a depth map may also introduce large errors. Therefore, our super-resolution network does not benefit the performance.


\section{Conclusion}
In this paper, we present a bundle adjustment framework to learn depth estimation from a single image using monocular videos. The  bundle adjustment framework uses photometric consistency to track pixels frame by frame, exploits depth consistency to connect distant frames, and introduces forward and backward motion to establish cross sequence constraints. Compared to previous algorithms, our algorithm generates more constraints and a larger baseline. We introduce a super-resolution network, which can produce a high resolution depth map  from a low resolution input. This solves the problem of interpolation which may result in large errors on object boundaries. Furthermore, we present the clip loss, which can make the training robust to moving objects and occlusions. The experimental results show that our algorithm is superior to the state-of-the-art unsupervised methods, and fills the gap between the methods using stereo and monocular training strategies.

{\small
\bibliographystyle{ieee}
\bibliography{deep_slam}
}

\end{document}